\documentclass[letterpaper]{article}

\usepackage{ijcai26}

\usepackage{times}
\usepackage{soul}
\usepackage{url}
\usepackage[small]{caption}
\usepackage{graphicx}
\usepackage{float}
\usepackage{amsmath}
\usepackage{amsthm}
\usepackage{booktabs}
\usepackage{algorithm}
\usepackage{algorithmic}
\usepackage[switch]{lineno}

\usepackage{amssymb}

\title{Fine-Tuning Diffusion Models for Molecular Generation via Reinforcement Learning and Fast Sampling}

\author{
Guang Lin$^1$
\and
Shikui Tu$^{1,*}$
\and
Lei Xu$^{1,2,*}$
\affiliations
$^1$Department of Computer Science and Engineering, Shanghai Jiao Tong University, Shanghai, China\\
$^2$Guangdong Laboratory of Artificial Intelligence and Digital Economy (SZ), Shenzhen, Guangdong, China
\emails
\{guanglin, tushikui, leixu\}@sjtu.edu.cn,
}

\begin{document}

\maketitle

\begin{abstract}
Generating molecules that simultaneously satisfy drug-like properties and conform to the 3D structure of a target protein is a core challenge in structure-based drug design (SBDD). Existing generative approaches, however, often rely on costly post-hoc processing during Sampling or require carefully curated datasets during training, yet still achieve modest gains. These limitations are especially pronounced in multi-objective settings, where balancing conflicting criteria remains a core challenge. To address these challenges, We propose FTDiff, a reinforcement learning fine-tuning framework tailored for diffusion-based molecular generation under structural constraints. To ensure stable and sample-efficient optimization, FTDiff adopts a group relative policy optimization (GRPO) style strategy. Furthermore, FTDiff builds upon a time-free pretrained diffusion model and incorporates a fast sampling mechanism that reduces the number of denoising steps, significantly accelerating both training and inference while maintaining generation quality. By optimizing a fixed threshold-aware reward, FTDiff effectively guides the model to produce valid, diverse, and high-quality molecules that balance multiple drug design objectives. Extensive experiments on benchmark datasets demonstrate that FTDiff consistently outperforms prior methods, without requiring expensive post-hoc optimization or intricate data engineering.
\end{abstract}

\section{Introduction}
\begin{figure}[!t]
\centering
\includegraphics[width=3.1in]{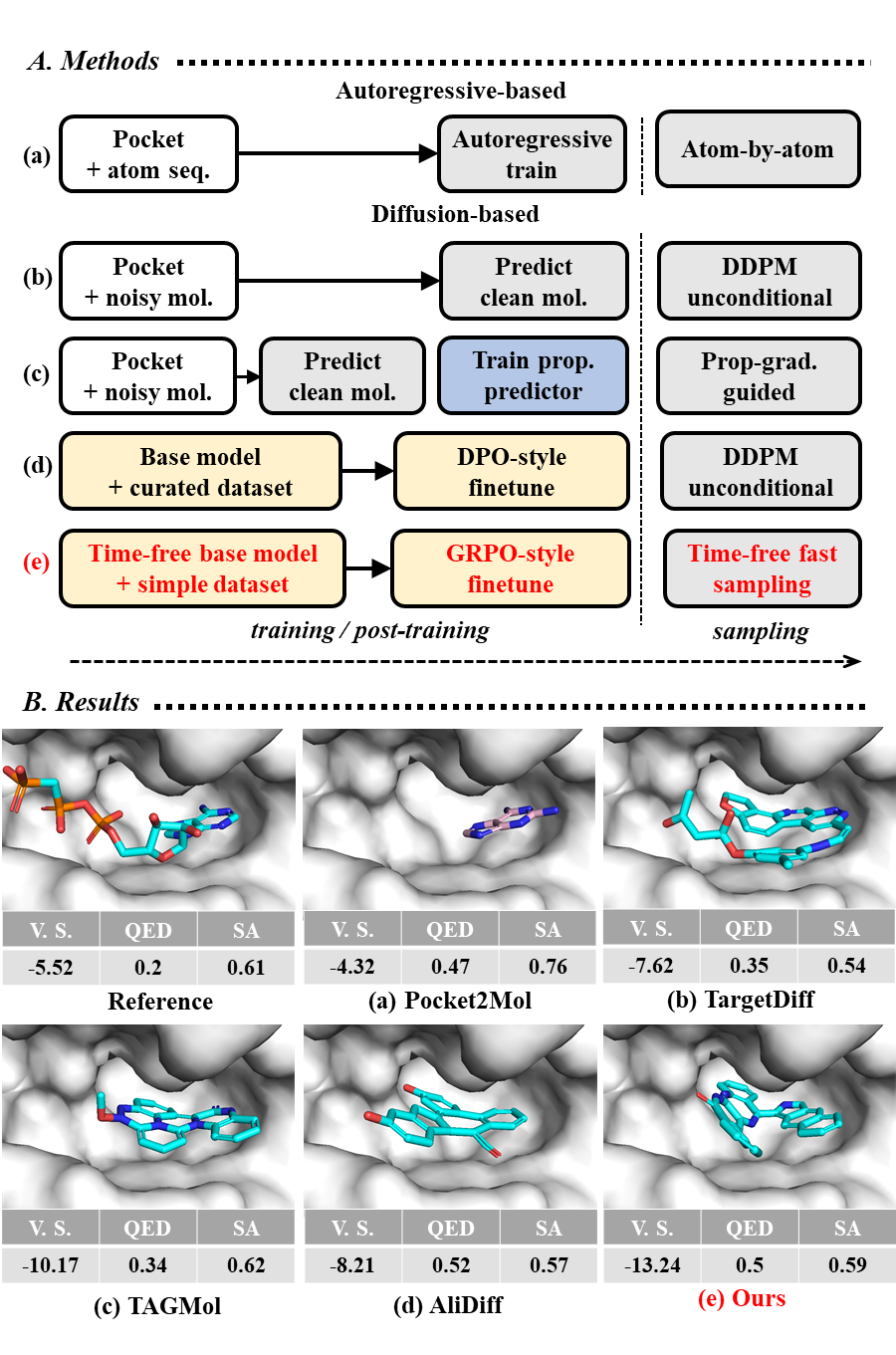} 
\caption{
\textbf{Comparison between our method and existing molecular generation approaches.}
\textbf{Top:} A taxonomy of molecular generation pipelines. Prior works typically follow autoregressive generation (a) Pocket2Mol or diffusion-based schemes, including plain generation (b) TargetDiff, property-guided optimization (c) TAGMol, and preference fine-tuning (d) ALIDiff.
\textbf{Bottom:} Visualization of generated molecules docked into a target protein pocket. Molecules generated by our method (e) achieve stronger binding (V.S., i.e., Vina Score) while maintaining favorable QED and SA values, demonstrating more effective multi-objective optimization.
}
\label{comparision_methods}
\end{figure}
Structure-based drug design (SBDD)\cite{anderson2003process} aims to generate molecules that can effectively bind to specific protein pockets. It plays a crucial role in modern drug discovery pipelines\cite{kalyaanamoorthy2011structure,lounnas2013current,scapin2006structural}, but remains a fundamentally challenging task due to the vastness of chemical space.  Generative models have recently emerged as a popular choice for this task~\cite{du2024machine}. By directly sampling molecules from learned distributions, generative models bypass the constraints of fixed libraries  which virtual screening suffers and allow for more efficient exploration of novel chemical structures. Depending on molecular representation, existing methods span from SMILES-based sequence models~\cite{grechishnikova2021transformer,qian2022alphadrug}, to 2D molecular graphs~\cite{tan2023target}, and more recently to 3D molecular structures~\cite{liu2022generating,peng2022pocket2mol,guan20233d,schneuing2024structure,zhou2024decompopt,qian2024kgdiff,dorna2024tagmol,gu2024aligning}. Among them, 3D-based approaches are particularly appealing for SBDD, as they naturally capture the spatial complementarity and geometric interactions between ligands and protein binding sites.

As illustrated in Fig~\ref{comparision_methods}, early 3D models often adopted autoregressive frameworks~\cite{liu2022generating,peng2022pocket2mol}, generating molecules atom-by-atom or fragment-by-fragment. While interpretable, these methods suffer from exposure bias and limited global structural optimization. To overcome these limitations, diffusion models have emerged as a powerful alternative for 3D molecular generation~\cite{guan20233d,zhou2024decompopt,gu2024aligning}, offering strong scalability and the ability to learn complex joint distributions over atom types and coordinates. Models like TargetDiff~\cite{guan20233d} and IPDiff~\cite{huang2024protein} formulate generation as SE(3)-equivariant denoising processes, often conditioned on protein's geometric priors.

While diffusion-based models have achieved promising results, their training objectives are typically limited to reconstructing molecules from noisy inputs. This formulation does not explicitly encourage optimization of pharmacologically relevant properties, and makes these models vulnerable to noise or bias in training datasets, resulting in suboptimal molecules. To improve property-specific outcomes, one line of work~\cite{qian2024kgdiff,dorna2024tagmol} introduces external guidance during sampling by computing gradients from auxiliary property predictors. Although this method allows direct optimization toward desired objectives, it significantly slows down sampling—especially under DDPM-style denoising due to repeated backward passes. Another approach~\cite{gu2024aligning} fine-tunes the diffusion model via direct preference optimization (DPO)~\cite{wallace2024diffusion}, but this requires carefully curated datasets that are costly to construct and may not generalize well. Both strategies face notable limitations in multi-objective scenarios, where optimizing one property often compromises others, and existing solutions yield only modest gains.

In this work, we introduce \textbf{FTDiff}, a multi-objective reinforcement learning framework for fine-tuning diffusion-based molecular generators. FTDiff efficiently adapts pretrained models toward structure-aware and property-optimized molecular design. It adopts a Group Relative Policy Optimization (GRPO)~\cite{shao2024deepseekmath} strategy that estimates advantages using intra-group reward statistics, eliminating the need for a learned value function. This design enables stable and sample-efficient optimization with low computational overhead. To further accelerate both training and inference, FTDiff builds on a time-free diffusion backbone that supports high-quality generation with as few as 10 denoising steps, achieving a significant speedup compared to standard DDPM-based methods~\cite{ho2020denoising}. For multi-objective optimization, FTDiff directly maximizes a composite reward that nonlinearly combines docking affinity, drug-likeness, and synthesizability, without relying on auxiliary property predictors. We evaluate FTDiff on both multi-objective and single-objective molecular design tasks, and demonstrate its superiority over representative and state-of-the-art baselines, generating diverse, valid, and high-affinity molecules that satisfy multiple drug-relevant criteria.

\noindent
\textbf{Our contributions are summarized as follows:}
\begin{itemize}
  \item We propose \textbf{FTDiff}, a fine-tuning framework for diffusion-based molecular generation that supports multiple objectives. It leverages reinforcement learning with a GRPO-style optimization strategy, enabling stable and efficient training.

  \item We introduce a fast sampling scheme based on time-free diffusion, reducing the denoising steps to 10 and significantly accelerating training and inference while preserving sample quality compared to 1000 steps in diffusion models.

  \item We design a simple yet effective threshold-aware reward that balances multiple drug design objectives, providing stable optimization signals for fine-tuning.

  \item We validate FTDiff on the CrossDocked2020~\cite{francoeur2020three} benchmark. Empirical results show that FTDiff achieves strong binding affinity with an average Vina score of $-7.18$, along with favorable molecular properties including Avg. QED of $0.56$ and Avg. SA of $0.61$, consistently outperforming competitive baselines.
\end{itemize}

\section{Method}
\begin{figure*}[!t]
\centering
\includegraphics[width=0.95\textwidth]{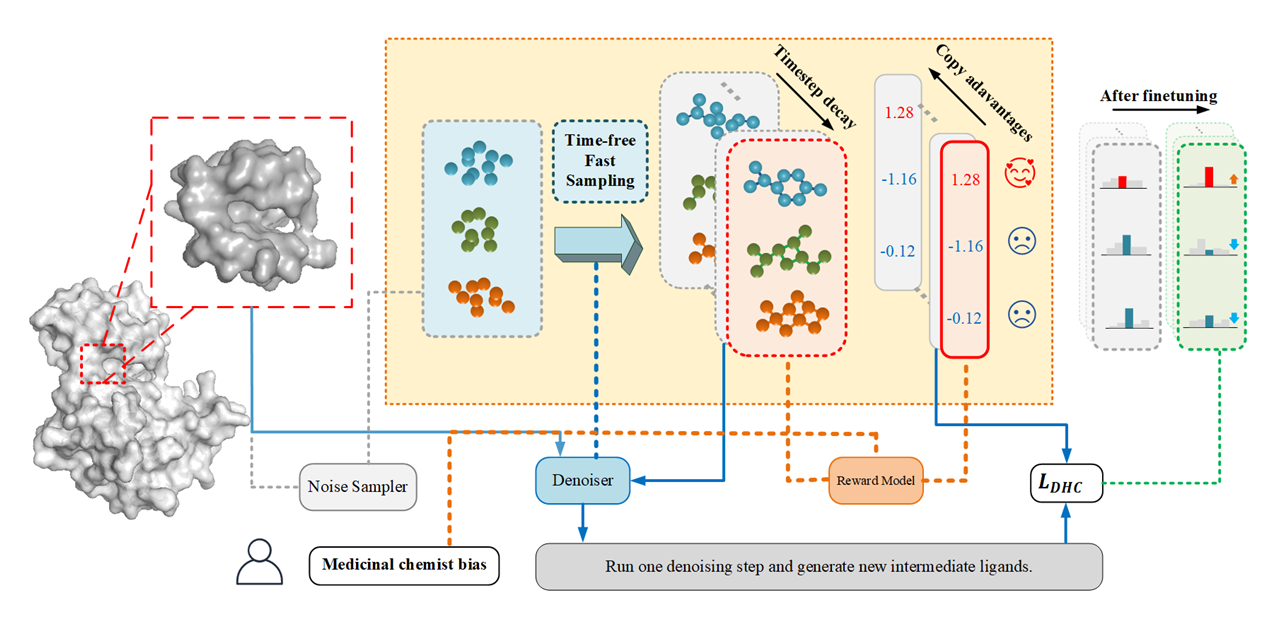}
\caption{
\textbf{Overview of the FTDiff fine-tuning framework.} 
For each protein pocket, we generate a batch of noisy molecules using an atom-count-preserving sampler. These noisy graphs are denoised via time-free fast sampling to produce full trajectories of intermediate states. The final molecule is evaluated using a multi-objective reward function, and the computed advantage is assigned to all states in the trajectory, which are then cached. During fine-tuning, cached intermediate states are reused to perform single-step denoising updates. Model parameters are optimized using a clipped double-head loss $L_{DHC}$ that compares new outputs with previous ones, guided by group-based advantage estimation for stable and efficient reward-driven learning.
}
\label{overview}
\end{figure*}
 
\subsection{FTDiff Overview}

As discussed in the previous section, our goal is to fine-tune a pretrained diffusion-based molecular generator to produce molecules that better satisfy multiple desired properties. We therefore propose \textbf{FTDiff}. As shown in Fig~\ref{overview}, \ the overall fine-tuning pipeline of FTDiff involves three stages: 
\textbf{1)} generating molecule candidates from a time-free pretrained diffusion model via fast sampling, \textbf{2)} evaluating these generated molecules using a multi-objective reward function, and \textbf{3)} updating the model based on reward differences to increase the probability of generating molecules that satisfy the desired objectives. A central challenge in fine-tuning diffusion models for molecular generation lies in achieving stable gradient updates and efficient sampling while effectively balancing multiple objectives. FTDiff addresses these issues by combining a GRPO-style policy optimization scheme that improves training stability,  a time-free fast sampling strategy that significantly reduces computational overhead, and a composite reward function that guides trade-offs among objectives, respectively. We elaborate on each component in the following sections.

\subsection{Policy Optimization Strategy}

A natural approach to fine-tuning molecular generators is to directly maximize the expected reward of the final generated molecule. Specifically, given a protein pocket $\mathcal{P}$ and a final generated molecule $\mathcal{M}_0 \sim p_\theta(\mathcal{M}_0 \mid \mathcal{P})$, one may attempt to optimize:
\begin{equation}
J(\theta) = \mathbb{E}_{\mathcal{P} \sim p(\mathcal{P}),\, \mathcal{M}_0 \sim p_\theta(\mathcal{M}_0 \mid \mathcal{P})} \left[ r(\mathcal{M}_0, \mathcal{P}) \right],
\end{equation}
where $r(\mathcal{M}_0, \mathcal{P})$ is a scalar reward function. However, this direct optimization overlooks the intrinsic multi-step denoising nature of diffusion models. Following the DDPO~\cite{black2023training}, we define the state as $s_t = (\mathcal{P}, t, \mathcal{M}_t)$, the action as $a_t = \mathcal{M}_{t-1}$, and the policy as $p_\theta(a_t \mid s_t) = p_\theta(\mathcal{M}_{t-1} \mid \mathcal{M}_t, \mathcal{P})$. The scalar reward is only assigned at the final step $t = 0$. Then, We collect denoising trajectories $\tau = \{\mathcal{M}_T, \mathcal{M}_{T-1}, \ldots, \mathcal{M}_0\}$ and apply Monte Carlo policy gradient:
\begin{equation}
\nabla_\theta J(\theta) \approx \mathbb{E}_\tau \left[ \sum_{t=1}^{T} \nabla_\theta \log p_\theta(\mathcal{M}_{t-1} \mid \mathcal{M}_t, \mathcal{P}) \cdot r(\mathcal{M}_0, \mathcal{P}) \right].
\end{equation}

To improve training stability and enable multiple optimization steps per batch, we adopt a GRPO-style~\cite{shao2024deepseekmath} policy optimization scheme with off-policy correction via importance sampling. Let $\theta_{\text{old}}$ be policy used to sample the trajectories the corrected gradient becomes:
\begin{align}
\nabla_\theta J(\theta) \approx 
\mathbb{E} \Bigg[ 
& \sum_{t=1}^{T} \sum_{i=1}^{G} w_t^{(i)} \cdot 
\nabla_\theta \log p_\theta(\mathcal{M}_{t-1}^{(i)} \mid 
\mathcal{M}_t^{(i)}, \mathcal{P}) \notag \\
& \cdot A_i(\mathcal{M}_0, \mathcal{P}) 
\Bigg],
\end{align}
where  $w_t^{(i)}$ is the importance ratio, $G$ denotes the group size and $A_i$ denotes the normalized advantage:

\begin{equation}
A_i(\mathcal{M}_0, \mathcal{P}) = \frac{r_i(\mathcal{M}_0, \mathcal{P}) - \mathrm{mean}(\{r_j\}_{j=1}^G)}{\mathrm{std}(\{r_j\}_{j=1}^G) + \epsilon}.
\end{equation}

We now furher decompose each molecule $\mathcal{M} = [\mathbf{x}, \mathbf{v}]$ into continuous atomic coordinates $\mathbf{x}$ and discrete atom types $\mathbf{v}$. The transition distribution of the diffusion model naturally factorizes as:
\begin{equation}
p_\theta(\mathcal{M}_{t-1} \mid \mathcal{M}_t, \mathcal{P}) = p_\theta(\mathbf{x}_{t-1} \mid \mathbf{x}_t, \mathcal{P}) \cdot p_\theta(\mathbf{v}_{t-1} \mid \mathbf{v}_t, \mathcal{P}).
\end{equation}
This factorization allows the importance ratio for off-policy correction to be written as:
\begin{equation}
w_t = \frac{p_\theta(\mathcal{M}_{t-1} \mid \mathcal{M}_t, \mathcal{P})}{p_{\theta_{\mathrm{old}}}(\mathcal{M}_{t-1} \mid \mathcal{M}_t, \mathcal{P})} = r_\mathbf{x} \cdot r_\mathbf{v},
\end{equation}
where
\begin{align}
r_\mathbf{x} &= \frac{p_\theta(\mathbf{x}_{t-1} \mid \mathbf{x}_t, \mathcal{P})}{p_{\theta_{\mathrm{old}}}(\mathbf{x}_{t-1} \mid \mathbf{x}_t, \mathcal{P})}, \\
r_\mathbf{v} &= \frac{p_\theta(\mathbf{v}_{t-1} \mid \mathbf{v}_t, \mathcal{P})}{p_{\theta_{\mathrm{old}}}(\mathbf{v}_{t-1} \mid \mathbf{v}_t, \mathcal{P})}.
\end{align}

While the full likelihood ratio $w_t = r_\mathbf{x} \cdot r_\mathbf{v}$ is mathematically correct under the factorized model, it assumes statistical independence between the coordinate prediction and atom-type prediction. However, in our conditional diffusion model, both $\mathbf{x}_{t-1}$ and $\mathbf{v}_{t-1}$ are predicted jointly by a shared neural network, conditioned on the same state $s_t = (\mathcal{M}_t, \mathcal{P})$. In practice, we approximate the policy optimization objective by treating $r_\mathbf{x}$ and $r_\mathbf{v}$ as partially disentangled components. Specifically, we compute the surrogate loss as the sum of two clipped terms, we named it as double head clipped loss:
\begin{equation}
\mathcal{L}_{\mathrm{DHC}} = \mathbb{E}_\tau \left[ \sum_{t=1}^T \left( \mathcal{L}_t^{\mathrm{clip}\text{-}\mathbf{x}} + \mathcal{L}_t^{\mathrm{clip}\text{-}\mathbf{v}} \right) \right],
\end{equation}
where
\begin{align}
\mathcal{L}_t^{\mathrm{clip}\text{-}\mathbf{x}} &= - \min \left( r_\mathbf{x} \cdot A, \operatorname{clip}(r_\mathbf{x}, 1-\epsilon, 1+\epsilon) \cdot A \right), \\
\mathcal{L}_t^{\mathrm{clip}\text{-}\mathbf{v}} &= - \min \left( r_\mathbf{v} \cdot A, \operatorname{clip}(r_\mathbf{v}, 1-\epsilon, 1+\epsilon) \cdot A \right).
\end{align}
Here, $\operatorname{clip}(\cdot)$ denotes the clipping operation, and $\epsilon$ controls the lower and upper bounds of clipping. If the gradients of individual coordinate components in $\mathcal{L}_t^{\mathrm{clip}\text{-}\mathbf{x}}$ are backpropagated separately, a rectified loss variant, denoted as $\mathcal{L}_{\mathrm{DHC}\_\mathrm{rectified}}$, is obtained. In $\mathcal{L}_{\mathrm{DHC}\_\mathrm{rectified}}$, the clipped coordinate loss is defined as
\begin{align}
\mathcal{L}_t^{\mathrm{clip}\text{-}\mathbf{x}}
=
-
\frac{1}{D}
\sum_{i=1}^{D}
\min \left(
r_{\mathbf{x}^{(i)}} \cdot A,\;
\operatorname{clip}(r_{\mathbf{x}^{(i)}}, 1-\epsilon, 1+\epsilon) \cdot A
\right),
\end{align}
where $D$ denotes the dimensionality of the spatial coordinates, and $r_{\mathbf{x}^{(i)}}$ is the probability ratio corresponding to the $i$-th coordinate component.


While it slightly relaxes the original PPO objective, it retains a trust-region constraint on both sub-policies and allows separate control over gradients from coordinate and type prediction.

\subsection{Time-Free Fast Sampling}


In diffusion-based generative models, accelerated sampling techniques such as DDIM~\cite{song2020denoising} have gained prominence due to their ability to reduce the number of sampling steps. These methods share a common idea: given a noisy input $\mathbf{z}_t$ and denoised predicton $\hat{\mathbf{z}}_0$, they construct a new sample $\mathbf{z}_{t-\Delta t}$ via resembling an interpolation between two endpoints.
The "interpolation" can equivalently be interpreted as structured noise reconstruction:
\begin{align}
    \mathbf{z}_{t-\Delta t} = s(t)\hat{\mathbf{z}}_0 &+  (s(t)\sigma(t) - \delta(t))\epsilon(\mathbf{z}_t, \hat{\mathbf{z}}_0) \notag \\
    &+ \sqrt{2\delta(t)(s(t)\sigma(t) - \delta(t))}\epsilon
\end{align}
Here, $s(t)$ denotes a time-dependent signal weighting factor, $\sigma(t)$ denotes the effective noise scale at time $t$, $\delta(t)$ is a small stochasticity parameter and $\epsilon \sim \mathcal{N}(0, I)$. However, these approaches are inherently dependent on explicit time-step scheduling, which becomes problematic in tasks such as molecular graph generation. In such setting, aggressive step skipping often leads to severe degradation of molecular validity and quality. We hypothesize that this is due to the difficulty of selecting appropriate time intervals $\Delta t$ that balance noise removal with structure preservation. Molecules are sensitive to minor structural distortions, and ill-timed denoising steps can easily result in broken bonds or chemically invalid outputs. 
 
To address this challenge, we propose a novel time-free adaptive denoising strategy. Our key insight is to remove the dependency on explicit time-step scheduling and instead learn to interpolate between $\hat{\mathbf{x}}_0$ and $\mathbf{x}_t$ using adaptive coefficients that are determined by the current state of the denoising process:



\begin{equation}
    \mathbf{x}_{t-\Delta t} = \text{scale} \cdot ( a(\hat{\mathbf{x}}_0,  {\mathbf{x}}_t)\hat{\mathbf{x}}_0 +  (1 - a(\hat{\mathbf{x}}_0,  {\mathbf{x}}_t)) {\mathbf{x}}_t),
\end{equation}
\begin{equation}
    \mathbf{h}_{t-\Delta t} = c(\hat{\mathbf{v}}_0, {\mathbf{v}_t})\hat{\mathbf{v}}_0 +  (1 - c(\hat{\mathbf{v}}_0, {\mathbf{v}_t})) {\mathbf{v}}_t,
\end{equation}
where
\begin{align}
     a(\hat{\mathbf{x}}_0,  {\mathbf{x}}_t) &= \frac{1}{1 + \gamma \text{log1p}(||\hat{\mathbf{x}}_0 - {\mathbf{x}}_t||_2^2)},\\
    c(\hat{\mathbf{v}}_0, {\mathbf{v}_t}) &= \frac{1}{1 +  \eta \text{log1p}[KL(\hat{\mathbf{v}}_0 || {\mathbf{v}_t})]}.
\end{align}
Here, the $|| \hat{\mathbf{x}}_0 - \mathbf{x}_t ||_2^2$ and the $\mathrm{KL}(\hat{\mathbf{v}}_0 || \mathbf{v}_t)$ serve as a surrogate for denoising confidence for two modalities, respectively.  By appropriately setting $\gamma$ and $\eta$, we encourage both modalities to progress through the denoising process in a synchronized manner.

\subsection{Reward Function Design}

A natural approach for multi-objective molecular optimization is to define the reward as a weighted sum of several normalized property scores:
\begin{equation}
r(\mathcal{M}_0, \mathcal{P}) = \sum_{i=1}^{n} w_i \cdot s_i,
\end{equation}
where $s_i \in [0, 1]$ represents the $i$-th normalized objective score, and $w_i$ is its corresponding weight. However, it treats all improvements as equally valuable, even beyond acceptability thresholds, potentially leading to over-optimization of certain objectives at the cost of others. To mitigate these issues, we propose a threshold-aware reward that applies a sigmoid transformation to each normalized score $s_i$, centered at a predefined threshold $\theta_i$. This yields $\sigma\left( \alpha (s_i - \theta_i) \right)$, where $\alpha$ controls sensitivity near the threshold—emphasizing improvements around critical acceptability regions. The overall reward is computed as:
\begin{equation}
r(\mathcal{M}_0, \mathcal{P}) = \sum_{i=1}^{n} w_i \cdot \sigma\left( \alpha (s_i - \theta_i) \right),
\end{equation}
where fixed weights $w_i$ reflects the relative importance of each objective.

\section{Experiments}
\subsection{Experimental Setup}


\subsubsection{Dataset}
We use the CrossDocked2020 dataset~\cite{francoeur2020three} with standard filtering~\cite{peng2022pocket2mol}, retaining ligand-pocket pairs with RMSD $<$ 1~\AA{} and sequence identity $<$ 30\%, yielding approximately 100k training samples. Evaluation is conducted on 100 held-out protein targets. For fine-tuning, we compute ESM-2~\cite{lin2022language} embeddings, apply K-means clustering, and select 100 representative pockets near cluster centers, ensuring diversity and efficiency. 

\subsubsection{Evaluation Metrics}

We evaluate the generated molecules in terms of binding affinity and key molecular properties. For binding affinity, we use four metrics: Vina Score, Vina Min, Vina Dock, and High Affinity. Vina Score estimates the binding affinity based on the generated 3D structures. Vina Min applies local minimization to refine the structures before scoring. Vina Dock performs re-docking to approximate the optimal binding configuration. High Affinity measures the proportion of generated molecules that outperform the reference ligand in terms of binding affinity for each protein target. We also evaluate key molecular properties by reporting the mean values of Quantitative Estimate of Drug-likeness (QED)\cite{bickerton2012quantifying}, Synthetic Accessibility (SA)\cite{ertl2009estimation}, and diversity.

\subsubsection{Reward Function and Sampling step}
In all experiments, we use a sigmoid threshold of $0.5$ with a sharpness coefficient $\alpha = 10$. The reward components, including normalized affinity, QED, and SA, are equally weighted for multi-objective optimization. During inference, we adopt a 10-step fast sampling scheme, achieving substantial speedup over the standard 1000-step process.

\subsubsection{Baseline}
We evaluate our proposed method, FTDiff, against a comprehensive set of baselines, which we broadly categorize into three groups: non-diffusion-based methods, diffusion-based methods, and diffusion-based methods enhanced with additional optimization complexity. The non-diffusion-based methods include autoregressive models such as  AR~\cite{luo20213d}, Pocket2Mol~\cite{peng2022pocket2mol}, and GraphBP~\cite{liu2022generating}, and  the Bayesian Flow Network model MolCRAFT~\cite{qu2024molcraft};  the diffusion-based methods include TargetDiff~\cite{guan20233d}, DecompDiff~\cite{guan2024decompdiff}, and IPDiff~\cite{huang2024protein}; the optimization-enhanced diffusion-based methods include  DecompOpt~\cite{zhou2024decompopt} with prior-guided search, TAGMol~\cite{dorna2024tagmol} employing gradient-based optimization, and ALIDiff~\cite{gu2024aligning} that applies DPO-style fine-tuning.


\begin{table*}[!htb]
\centering
\resizebox{\textwidth}{!}{
\begin{tabular}{clrrrrrrrrrrr}
\toprule
    \multicolumn{2}{c}{\textbf{Methods}} & \multicolumn{2}{r}{Vina Score ($\downarrow$)} & \multicolumn{2}{r}{Vina Min ($\downarrow$)} & \multicolumn{2}{r}{Vina Dock ($\downarrow$)} & \multicolumn{2}{r}{High Affinity ($\downarrow$)} & QED ($\uparrow$) & SA ($\uparrow$) & Div. ($\uparrow$)\\
    & & Avg. & Med. & Avg. & Med. & Avg. & Med. & Avg. & Med. & Avg. & Avg. & Avg.\\
\midrule
    \multicolumn{2}{c}{\textbf{Reference}} & -6.36 & -6.46 & -6.71 & -6.49 & -7.45 & -7.26 & - & - & 0.48 & 0.73 & - \\
\midrule
     \textbf{Comp. w/ } &
    LiGAN & -     & -     & -     & -  & -6.33 & -6.20 & 21.1\% & 11.1\% & 0.39  &0.59 & 0.66\\
    \textbf{Non-diff.} &  GraphBP & -     & -     & -     & -     & -4.80 & -4.70 & 17.2\% & 6.7\% & 0.43 & 0.49 & \textbf{0.79}\\
    & AR & -5.75 & -5.64 & -6.18 & -5.88 & -6.75 & -6.62 & 37.9\% & 31.0\% & 0.51 & 0.63 & 0.70 \\

    & Pocket2Mol & -5.14 & -4.70 & -6.42 & -5.82 & -7.15 & -6.79 & 48.4\% & 51.0\% & \underline{0.56} & \textbf{0.74} & 0.69 \\
    & MolCRAFT  & \underline{-6.59} & \underline{-7.04} & \underline{-7.27} & \underline{-7.26} & \underline{-7.92} & \underline{-8.01} & \underline{59.1\%}& \underline{62.6\%} & 0.50 & \underline{0.68} & 0.72 \\
    & FTDiff & \textbf{-7.18} & \textbf{-7.73}& \textbf{-8.48}& \textbf{-8.53} &  \textbf{-9.44} & \textbf{-9.54} & \textbf{78.6\%} & \textbf{88.9\%} & \textbf{0.56} & 0.61 & \underline{0.72}\\
\midrule
    \textbf{Comp. w/}
        & TargetDiff & -5.47 & -6.30 & -6.64 & -6.83 & -7.8 & -7.91 & 58.1\% & 59.1\% & 0.48 & 0.58 & 0.72 \\
    \textbf{ diff.} & DecompDiff & -5.67 & -6.04 & -7.04 & -7.09 & -8.39 & -8.43 & 64.4\% & 71.0\% & 0.45 & 0.61 & 0.68 \\
    & IPDiff & \underline{-6.42} & \underline{-7.01} & \underline{-7.45} & \underline{-7.48} & \underline{-8.57} & \underline{-8.51} & \underline{69.5\%} & \underline{75.5\%} & \underline{0.52} & \underline{0.61} & \textbf{0.74} \\
    & FTDiff & \textbf{-7.18} & \textbf{-7.73}& \textbf{-8.48}& \textbf{-8.53} &  \textbf{-9.44} & \textbf{-9.54} & \textbf{78.6\%} & \textbf{88.9\%} & \textbf{0.56} & \textbf{0.61} & \underline{0.72}\\
\midrule
    \textbf{Comp. w/ }
    & DecompOpt  & -5.87 & -6.81 & -7.35 & -7.72 & \underline{-8.98} & \underline{-9.01} & \underline{73.5\%} & \textbf{93.3\%} & 0.48 & \textbf{0.65} & 0.60 \\
    \textbf{diff.+opt.}& TAGMol  & -7.02 & \underline{-7.77} & -7.95 & -8.07 & -8.59 & -8.69& 69.8\% & 76.4\% & \underline{0.55} & 0.56 & 0.69\\
    & ALIDiff  & \underline{-7.07} & \textbf{-7.95}& \underline{-8.09}& \underline{-8.17} & -8.9 & -8.81 & 73.4\%& 81.4\% & 0.5 & 0.57 & \textbf{0.73}\\
    & FTDiff & \textbf{-7.18} & -7.73 & \textbf{-8.48}& \textbf{-8.53} &  \textbf{-9.44} & \textbf{-9.54} & \textbf{78.6\%} & \underline{88.9\%} & \textbf{0.56} & \underline{0.61} & \underline{0.72}\\
\bottomrule
\end{tabular}}
\caption{Summary of binding affinity and molecular properties of reference molecules and molecules generated by baselines and FTDiff under the multi-objective setting. ($\uparrow$) / ($\downarrow$) denotes whether a larger / smaller number is preferred.
Top 2 results are bolded and underlined, respectively.}
\label{tab:main_results}
\end{table*}

\subsection{Comparison on Multi-Objective Optimization}

We present a comprehensive comparison of FTDiff against a wide range of baselines under a multi-objective optimization setting. As summarized in Table~\ref{tab:main_results}, FTDiff consistently achieves the best or second-best performance across all major evaluation metrics.

\subsubsection{Superior Binding Capability.} 
FTDiff achieves the strongest binding affinity among all evaluated methods. It attains the best performance across all docking-related metrics, including an average Vina Score of -7.18, Vina Min of -8.48, and Vina Dock of -9.44. Compared to the best-performing non-diffusion baseline, MolCRAFT, FTDiff improves the Avg. Vina Score by 9.0\%, Vina Min by 16.6\%, and Vina Dock by 19.2\%. Against the strongest diffusion-based baseline, IPDiff, FTDiff achieves relative gains of 11.8\%, 13.8\%, and 10.9\%, respectively. Even compared with optimization-based models, FTDiff consistently outperforms all baselines, demonstrating that reinforcement learning-based fine-tuning enables more effective alignment between molecular generation and docking objectives.

\subsubsection{Higher Success Rate in Binding Optimization.}
FTDiff generates a significantly larger fraction of high-affinity molecules compared to all baselines. It achieves a mean high-affinity rate of 78.6\% and a median of 88.9\%, outperforming non-optimization models such as IPDiff with 69.5\% and MolCRAFT with 59.1\%, as well as optimization-based pipelines including ALIDiff at 73.4\% and TAGMol at 69.8\%. While DecompOpt attains a higher median rate of 93.3\%, its lower mean of 73.5\% indicates that FTDiff offers a better balance of strength and stability across sampled molecules.

\subsubsection{Balanced Drug-likeness and Synthesizability.}
FTDiff achieves a strong balance between molecular quality and synthetic accessibility. It attains the highest QED score of 0.56, matching the best baseline Pocket2Mol, and maintains a competitive SA score of 0.61. Although MolCRAFT achieves a slightly higher SA score of 0.74, its performance on binding metrics is weaker. In contrast, FTDiff combines high drug-likeness and synthesizability with superior binding affinity, highlighting its ability to generate molecules that are both potent and practically viable.

\subsubsection{Maintaining Diversity Under Optimization.}
FTDiff preserves high molecular diversity under reward-driven fine-tuning, achieving a score of 0.72. It ranks closely behind GraphBP and IPDiff, and outperforms most optimization-based methods such as DecompOpt and TAGMol. This demonstrates that sampling-based reinforcement learning promotes both structural diversity and alignment with task objectives.

\subsection{Robustness Under Multi-Objective Constraints}

\begin{table}[!htb]
\centering
\resizebox{0.48\textwidth}{!}{
\begin{tabular}{lrrrr}
\toprule
Methods & Aff.$<$-8. & + QED$>$0.5 & +SA$>$0.5  & Con. \\
\midrule
AR            & 20.3 & 12.4 & 7.7  & 93.0\\
Pocket2Mol    & 16.2 & 8.9  & 8.4  & \textbf{98.3} \\
MolCRAFT     & 36.2 & 17.5 & 16.7 & \underline{96.7}\\
TargetDiff     & 26.3 & 12.5 & 8.6 & 90.4\\
TAGMol        & \underline{47.0} & \underline{27.9} & \underline{17.4} & 92.0 \\
ALIDiff        & 46.6 & 21.3 & 14.1 & 92.4\\
FTDiff         & \textbf{47.4} & \textbf{30.4} & \textbf{22.3} & 95.8\\
\bottomrule
\end{tabular}
}
\caption{Pass rates (\%) of generated molecules under increasingly strict criteria: 
(1) Vina score $<$ -8; (2) Vina score $<$ -8 and QED $>$ 0.5; 
(3) Vina score $<$ -8, QED $>$ 0.5, and SA $>$ 0.5. ‘Con.’ denotes molecules with no broken bonds. Top-2 results are \textbf{bolded} and \underline{underlined}.}
\label{tab:pass_rate_results}
\end{table}

To assess the robustness under realistic multi-objective requirements, we evaluate the percentage of generated molecules that satisfy increasingly stringent thresholds across three key metrics: Vina score $<$ -8, QED $>$ 0.5, and SA $>$ 0.5. As shown in Table~\ref{tab:pass_rate_results}, FTDiff consistently outperforms all baselines. It achieves a pass rate of 47.4\% for Vina score below $-8$, comparable to the best-performing methods TAGMol and ALIDiff. With the added QED constraint, FTDiff reaches 30.4\%, exceeding TAGMol by 2.5 points. Under the strictest joint constraint, FTDiff achieves 22.3\%, outperforming the next-best method by 8.2 points. These results suggest that while many models excel at affinity optimization, they often neglect molecular quality or feasibility. In contrast, FTDiff’s reward-guided fine-tuning balances all objectives, yielding valid, drug-like, and synthesizable molecules. Notably, it also maintains a high connectivity rate of 95.8\%, reinforcing its practical utility in structure-based drug design.

\subsection{Comparison with Single-Objective Optimization Methods}

\begin{table*}[!htbp]
\centering
\begin{tabular}{lrrrrrrrrrrr}
\toprule
    \multicolumn{1}{c|}{\textbf{Methods}} & \multicolumn{2}{c|}{Vina Score ($\downarrow$)} & \multicolumn{2}{c|}{Vina Min ($\downarrow$)} & \multicolumn{2}{c|}{Vina Dock ($\downarrow$)} & \multicolumn{2}{c|}{High Affinity ($\downarrow$)} & QED ($\uparrow$) & SA ($\uparrow$) & Div. ($\uparrow$)\\
    & Avg. & Med. & Avg. & Med. & Avg. & Med. & Avg. & Med. & Avg. & Avg. & Avg. \\
\midrule
    Reference & -6.36 & -6.46 & -6.71 & -6.49 & -7.45 & -7.26 & - & - & 0.48 & \textbf{0.73} & - \\
    KGDiff & -8.04 & \textbf{-8.61} & -8.78 & -8.85 & -9.43 & -9.43 & 79.2\% & 87.0\% & \textbf{0.51} & 0.54 & 0.75 \\
    FTDiff & \textbf{-8.11} & -8.25 & \textbf{-9.41} & \textbf{-9.28} & \textbf{-10.22} & \textbf{-10.09} & \textbf{80.7\%} & \textbf{89\%} & 0.49 & 0.53 & \textbf{0.75} \\
\bottomrule
\end{tabular}
\caption{Summary of binding affinity and molecular properties of reference molecules and molecules generated by baselines and FTDiff under the single-objective setting. ($\uparrow$) / ($\downarrow$) denotes whether a larger / smaller number is preferred.}
\label{tab:single_results}
\end{table*}

We further compare FTDiff with KGDiff, a state-of=art method designed for optimizing binding affinity through gradient-guided generation. KGDiff jointly predicts molecular affinity scores during training and leverages gradients to guide atomic coordinates and types at sampling time. This direct optimization mechanism yields strong performance on affinity-oriented metrics.
As shown in Table~\ref{tab:single_results}, both KGDiff and FTDiff substantially outperform the reference ligands across all affinity metrics. Notably, FTDiff achieves the best overall performance in all evaluated affinity indicators. It surpasses KGDiff by a margin of 0.07 on average Vina Score, 0.63 on Vina Min, and 0.79 on Vina Dock, indicating stronger binding potential and better docking poses. Moreover, FTDiff achieves the highest High Affinity Rate of 89\%, compared to 87.0\% by KGDiff, reflecting its robustness across targets.

\subsection{More discussion}

\subsubsection{Effectiveness of our approach} 

\begin{table*}[!htbp]
\centering
\resizebox{\textwidth}{!}{
\begin{tabular}{lrrrrrrrrrrr}
\toprule
    \multicolumn{1}{c}{\textbf{Methods}} & \multicolumn{2}{c}{Vina Score ($\downarrow$)} & \multicolumn{2}{c}{Vina Min ($\downarrow$)} & \multicolumn{2}{c}{Vina Dock ($\downarrow$)} & \multicolumn{2}{c}{High Affinity ($\downarrow$)} & QED ($\uparrow$) & SA ($\uparrow$) & Div. ($\uparrow$) \\
    & Avg. & Med. & Avg. & Med. & Avg. & Med. & Avg. & Med. & Avg. & Avg. & Avg. \\
\midrule
    Pretrained & -6.39 & -6.82 & -7.71 & -7.64 & -8.55 & -8.52 & 65.6\% & 72.7\% & 0.48 & 0.55 & 0.68 \\
    FTDiff-$L_{\mathrm{Vanilla}}$ & -6.45 & -6.53 & -8.01 & -7.78 & -9.01 & -8.92 & 74.8\% & 83.5\% & 0.57 & 0.59 & \textbf{0.73} \\
    FTDiff-$L_{\mathrm{DHC}}$ & -6.29 & -6.96 & -8.18 & -8.32 & -9.11 & -9.23 & 74.3\% & 81.8\% & \textbf{0.57} & 0.58 & 0.71 \\
    FTDiff-$L_{\mathrm{DHC}\_\mathrm{rectified}}$ & \textbf{-7.18} & \textbf{-7.73} & \textbf{-8.48} & \textbf{-8.53} & \textbf{-9.44} & \textbf{-9.54} & \textbf{78.6\%} & \textbf{88.9\%} & 0.56 & \textbf{0.61} & 0.72 \\
\bottomrule
\end{tabular}}
\caption{Effect of reinforcement‑learning fine‑tuning (FTDiff) on binding affinity and molecular properties. ($\uparrow$) / ($\downarrow$) denotes whether a larger / smaller number is preferred.}
\label{tab:effective_Of_finetune}
\end{table*}

To assess the effect of fine-tuning, we compare FTDiff with its pretrained counterpart. As shown in Table~\ref{tab:effective_Of_finetune}, FTDiff improves the Avg. Vina score by 12.4\%  and increases the proportion of high-affinity molecules by 19.8\%. Besides affinity, FTDiff preserves molecular quality with a 16.7\% gain in QED and an 11.0\% improvement in SA. Diversity also increases by 5.9\%, suggesting enhanced chemical exploration post fine-tuning. Overall, these results confirm that reinforcement learning with reward-guided optimization not only strengthens target-specific binding but also retains or improves essential drug-like properties.

\subsubsection{Comparision of Fine-Tuning Objectives}

\begin{figure}[!htb]
\centering
\includegraphics[width=3.2in]{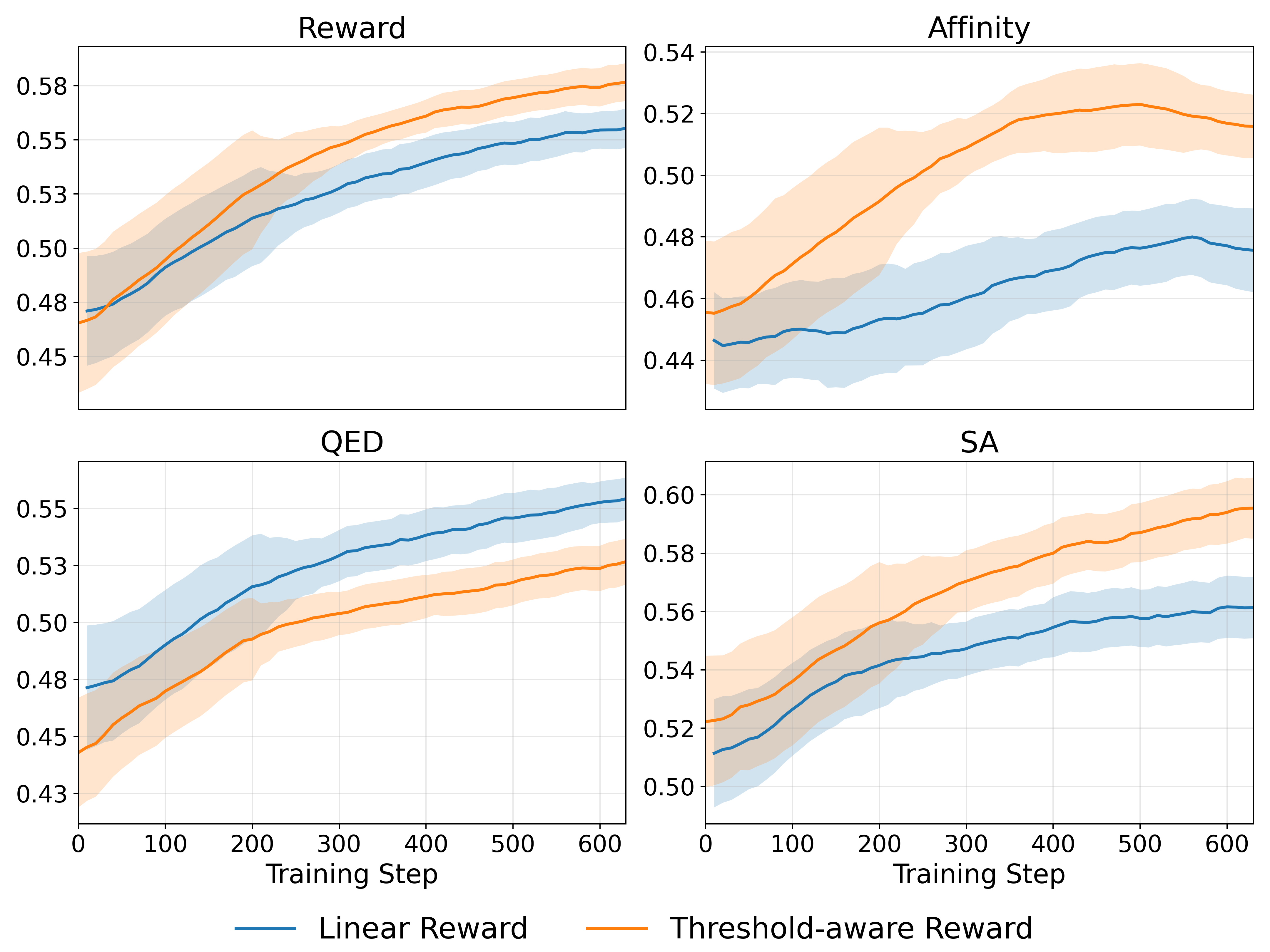}
    \caption{
        Effect of different reward function designs on molecular generation training curves.
The left block shows composite rewards computed by linear weighting (Linear Reward), while the right block shows threshold-aware Sigmoid rewards (Threshold-aware Reward). Each curve represents the performance of a method, and the shaded area indicates the variability (standard deviation) during training.
    }
    \label{fig:ablation_test_subset}
\end{figure}

To analyze the impact of different fine-tuning objectives, we compared the standard denoising loss $L_{\mathrm{vanilla}}$ with the two proposed variants of $L_{\mathrm{DHC}}$. As shown in Figure~\ref{fig:ablation_test_subset}, $L_{\mathrm{DHC}\_\mathrm{rectified}}$ significantly outperforms $L_{\mathrm{vanilla}}$ in terms of reward and combined affinity, while maintaining comparable performance on QED and SA. Notably, $L_{\mathrm{vanilla}}$ struggles to achieve consistent improvement in affinity, whereas $L_{\mathrm{DHC}\_\mathrm{rectified}}$ provides stable and substantial gains.

\subsubsection{Comparison of Reward Functions}

\begin{figure}[!htb]
\centering
\includegraphics[width=3.2in]{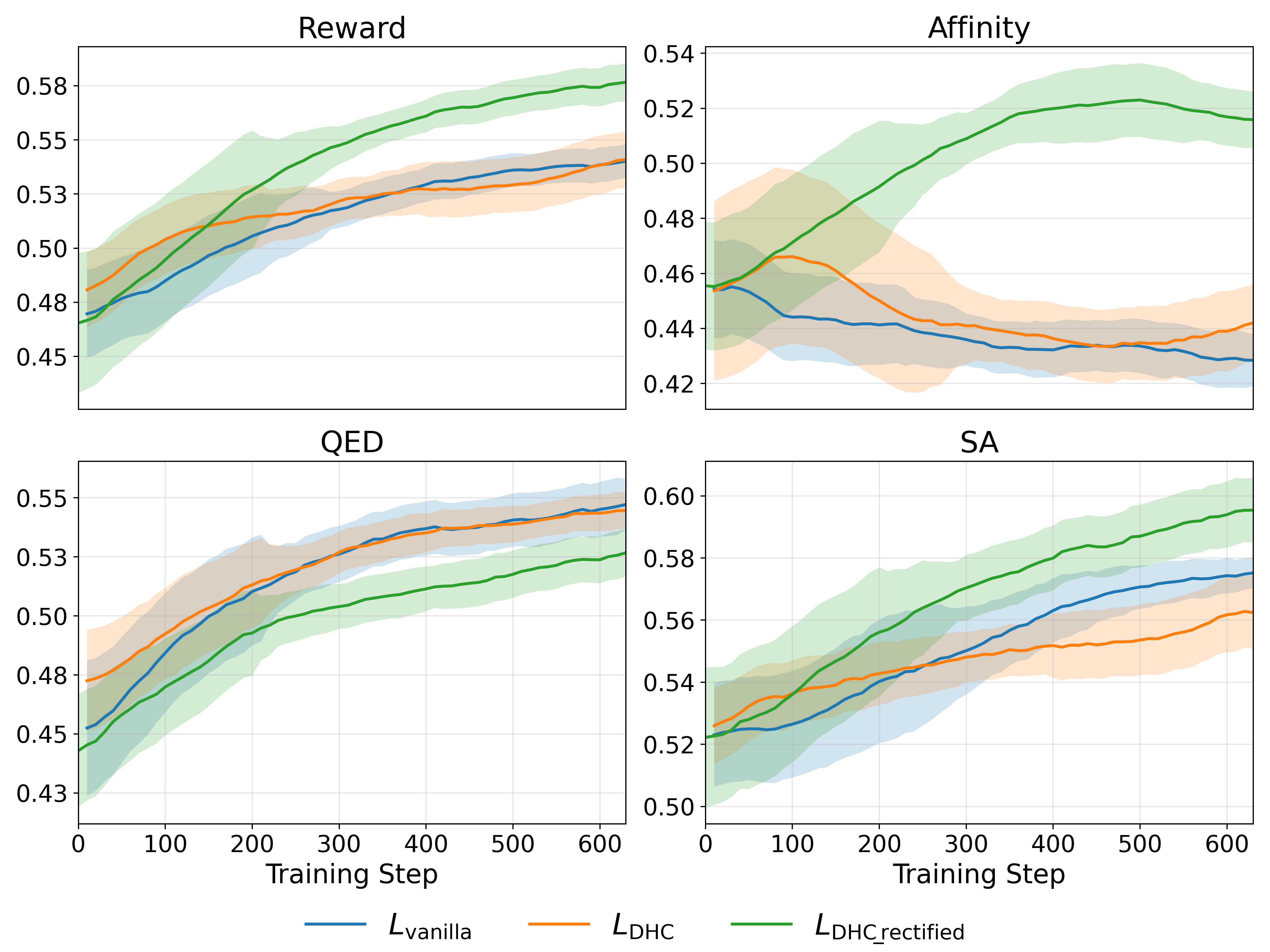}
    \caption{
        Evaluation on the held-out test subset during fine-tuning. 
        This figure compares the performance of models fine-tuned with two different objectives: 
        the standard loss $L_{\mathrm{vanilla}}$ and our proposed double head clipped loss $L_{\mathrm{DHC}}$. 
        Four evaluation metrics are plotted over training steps: reward, affinity score, QED, and SA. 
        All curves are computed on a held-out subset of the test set to assess generalization behavior during training.
    }
    \label{fig:ablation_test_rewards}
\end{figure}

As shown in Figure~\ref{fig:ablation_test_rewards}, we compare the performance of two composite reward functions in the multi-objective molecular generation task. When using a linear weighting scheme, the model performs poorly in terms of both affinity and SA, although it achieves slightly higher scores on QED. In contrast, our proposed threshold-aware reward function more effectively guides the model to optimize around key property metrics, enabling all metrics to exceed the specified 0.5 threshold while avoiding excessive optimization of QED. This results in a significant improvement in overall performance.

\section{Conclusion}

In this work, we present \textbf{FTDiff}, a novel framework for fine-tuning pretrained diffusion models toward generating molecules that better satisfy multiple complex objectives. By integrating a GRPO-style policy optimization scheme, FTDiff achieves stable and effective gradient updates despite the inherent challenges of multi-step denoising processes. Our time-free fast sampling reduces denoising steps from 1000 to 10, yielding a substantial acceleration in inference without compromising generation quality, effectively overcoming a major bottleneck in diffusion-based molecular design.  Extensive experiments on the CrossDocked2020 demonstrate that FTDiff consistently improves molecular quality across various evaluation metrics. Our method shows strong performance in both single-objective and multi-objective settings, effectively enhancing binding affinity while maintaining drug-likeness and synthesizability. While FTDiff demonstrates promising results, there remains room for improvement. One potential direction is to incorporate dynamic weighting schemes during fine-tuning to better explore the Pareto front of multiple objectives. This would allow the model to adaptively balance trade-offs across diverse property combinations, enabling broader applicability in real-world multi-objective drug design.

\bibliographystyle{named}
\bibliography{ijcai26}

\clearpage
\appendix

\section*{Ethical Statement}

There are no ethical issues.


\section{Equivariant Conditional Diffusion Modeling}

We adopt an equivariant conditional diffusion framework for structure-based
molecular generation, following prior work ~\cite{guan20233d}.
A ligand molecule is modeled
as a continuous random variable composed of atom types and three-dimensional coordinates, and its conditional distribution is learned via a diffusion process.

The protein binding pocket is represented as a set of atoms
$\mathcal{P} = \{(\mathbf{x}_P^{(i)}, \mathbf{v}_P^{(i)})\}_{i=1}^{N_P}$,
where $N_P$ denotes the number of protein atoms,
$\mathbf{x}_P^{(i)} \in \mathbb{R}^3$ denotes the three-dimensional coordinate of
the $i$-th atom, and
$\mathbf{v}_P^{(i)} \in \mathbb{R}^{N_f}$ represents atom-level features such as
element type and amino acid identity.

A ligand molecule is represented as
$\mathcal{M} = \{(\mathbf{x}_L^{(i)}, \mathbf{v}_L^{(i)})\}_{i=1}^{N_M}$,
which is compactly written as $\mathcal{M} = [\mathbf{x}, \mathbf{v}]$,
where $[\cdot, \cdot]$ denotes vector concatenation,
$\mathbf{x} \in \mathbb{R}^{N_M \times 3}$ represents atomic coordinates,
and $\mathbf{v} \in \mathbb{R}^{N_M \times K}$ is a one-hot encoding of atom types.

The forward diffusion process is defined over both atomic coordinates and atom
types.
Continuous coordinates $\mathbf{x}$ are modeled with Gaussian transitions,
while discrete atom types $\mathbf{v}$ are modeled with categorical transitions.
The joint molecular distribution is factorized into the product of these two
components.
Specifically, given a fixed variance schedule
$\beta_1, \dots, \beta_T$, the forward process at timestep $t$ is defined as:
\begin{equation}
\begin{aligned}
q(\mathcal{M}_t \mid \mathcal{M}_{t-1}, \mathcal{P}) =
\mathcal{N}\big(\mathbf{x}_t; \sqrt{1-\beta_t}\,\mathbf{x}_{t-1}, \beta_t I\big)
\\ \cdot
\mathcal{C}\big(\mathbf{v}_t \mid (1-\beta_t)\mathbf{v}_{t-1} + \beta_t / K \big),
\end{aligned}
\end{equation}
where $K$ denotes the number of atom types.

The marginal distribution at an arbitrary timestep admits a closed-form
expression:
\begin{align}
q(\mathbf{x}_t \mid \mathbf{x}_0) &=
\mathcal{N}(\mathbf{x}_t; \sqrt{\bar{\alpha}_t}\,\mathbf{x}_0,
(1-\bar{\alpha}_t) I), \\
q(\mathbf{v}_t \mid \mathbf{v}_0) &=
\mathcal{C}\big(\mathbf{v}_t \mid
\bar{\alpha}_t \mathbf{v}_0 + (1-\bar{\alpha}_t)/K\big),
\end{align}
where $\bar{\alpha}_t = \prod_{s=1}^t (1-\beta_s)$.

By Bayes' rule, the posterior distributions over atomic coordinates and atom
types can be derived in closed form:
\begin{align}
q(\mathbf{x}_{t-1} \mid \mathbf{x}_t, \mathbf{x}_0) &=
\mathcal{N}(\mathbf{x}_{t-1};
\tilde{\mu}_t(\mathbf{x}_t, \mathbf{x}_0), \tilde{\beta}_t I), \\
q(\mathbf{v}_{t-1} \mid \mathbf{v}_t, \mathbf{v}_0) &=
\mathcal{C}(\mathbf{v}_{t-1} \mid \tilde{c}_t(\mathbf{v}_t, \mathbf{v}_0)),
\end{align}
with
\begin{align}
\tilde{\mu}_t(\mathbf{x}_t, \mathbf{x}_0) &=
\frac{\sqrt{\bar{\alpha}_{t-1}} \beta_t}{1-\bar{\alpha}_t} \mathbf{x}_0
+
\frac{\sqrt{\alpha_t}(1-\bar{\alpha}_{t-1})}{1-\bar{\alpha}_t} \mathbf{x}_t, \\
\tilde{\beta}_t &= \frac{1-\bar{\alpha}_{t-1}}{1-\bar{\alpha}_t} \beta_t, \\
\tilde{c}_t(\mathbf{v}_t, \mathbf{v}_0) &=
\frac{c^\star(\mathbf{v}_t, \mathbf{v}_0)}
{\sum_{k=1}^K c_k^\star(\mathbf{v}_t, \mathbf{v}_0)}, \\
c^\star(\mathbf{v}_t, \mathbf{v}_0) &=
\big[\alpha_t \mathbf{v}_t + (1-\alpha_t)/K\big]
\odot
\big[\bar{\alpha}_{t-1} \mathbf{v}_0 + (1-\bar{\alpha}_{t-1})/K\big].
\end{align}

The reverse generative process progressively transforms an initial noise sample
$\mathcal{M}_T$ into a molecular structure $\mathcal{M}_0$.
The reverse transition is parameterized by a neural network:
\begin{equation}
\begin{aligned}
p_\theta(\mathcal{M}_{t-1} \mid \mathcal{M}_t, \mathcal{P}) =
\mathcal{N}\big(\mathbf{x}_{t-1};
\mu_\theta([\mathbf{x}_t, \mathbf{v}_t], t, \mathcal{P}), \sigma_t^2 I\big)\\
\cdot
\mathcal{C}\big(\mathbf{v}_{t-1} \mid
c_\theta([\mathbf{x}_t, \mathbf{v}_t], t, \mathcal{P})\big).
\end{aligned}
\end{equation}

To model protein--ligand spatial interactions, we employ an SE(3)-equivariant
graph neural network that jointly encodes ligand atoms and protein pocket
residues.
Protein structural information is incorporated as conditioning signals during
denoising, enabling the generation of three-dimensional ligand structures that
satisfy both geometric constraints and chemical validity.

\section{Derivation of Time-Free Fast Sampling}
\subsection{Noise Scheduling}
We first sample time steps using the Symmetric Sampling Method
\[
    t \sim \mathcal{U}(0, T) 
\]
and then calculate the noise level. Due to the need to perturb two different aspects, namely atoms' positions and types of ligands, we need to set up two types of scheduling. 

\begin{enumerate}
\item \textbf{Positions' scheduling:} we set $\beta_{start}=1.e-7$, $\beta_{end}=0.002$, and 
\[
    \beta_t= \text{sigmoid}\Big(-6 + \frac{12(t - 1)}{ T - 1}\Big)\cdot(\beta_{\text{end}} - \beta_{\text{start}}) + \beta_{\text{start}}
\]
and we have
\[
    \alpha_t = 1 - \beta_t, \quad
    \overline{\alpha}_t = \sum_{s=0}^t \alpha_s
\]
Therefore, we obtain $\alpha_t$ and $\overline{\alpha}_t$ from $t$.

\item \textbf{Types' scheduling:} we set $s=0.01$. Here, we use the symbol $\beta$ to denote the scheduling parameter, which corresponds to $\alpha$ in Section~A. 
The change in notation is intended to clearly distinguish the noise scheduling for atom types from that for coordinates:
\[
    \overline{\beta}_t = \left( \cos\left( \frac{t/T + s}{1 + s} \cdot \frac{\pi}{2} \right) \right)^2
\]
\[
    \beta_t = \sqrt{\text{clip}\Big(\frac{\overline{\beta}_{t+1}}{\overline{\beta}_t}, 0.001, 1.0\Big)}
\]
\end{enumerate}

\subsection{Time-Free Fast Sampling Interpolation}

\begin{figure*}[!t]
\centering
\includegraphics[width=0.95\textwidth]{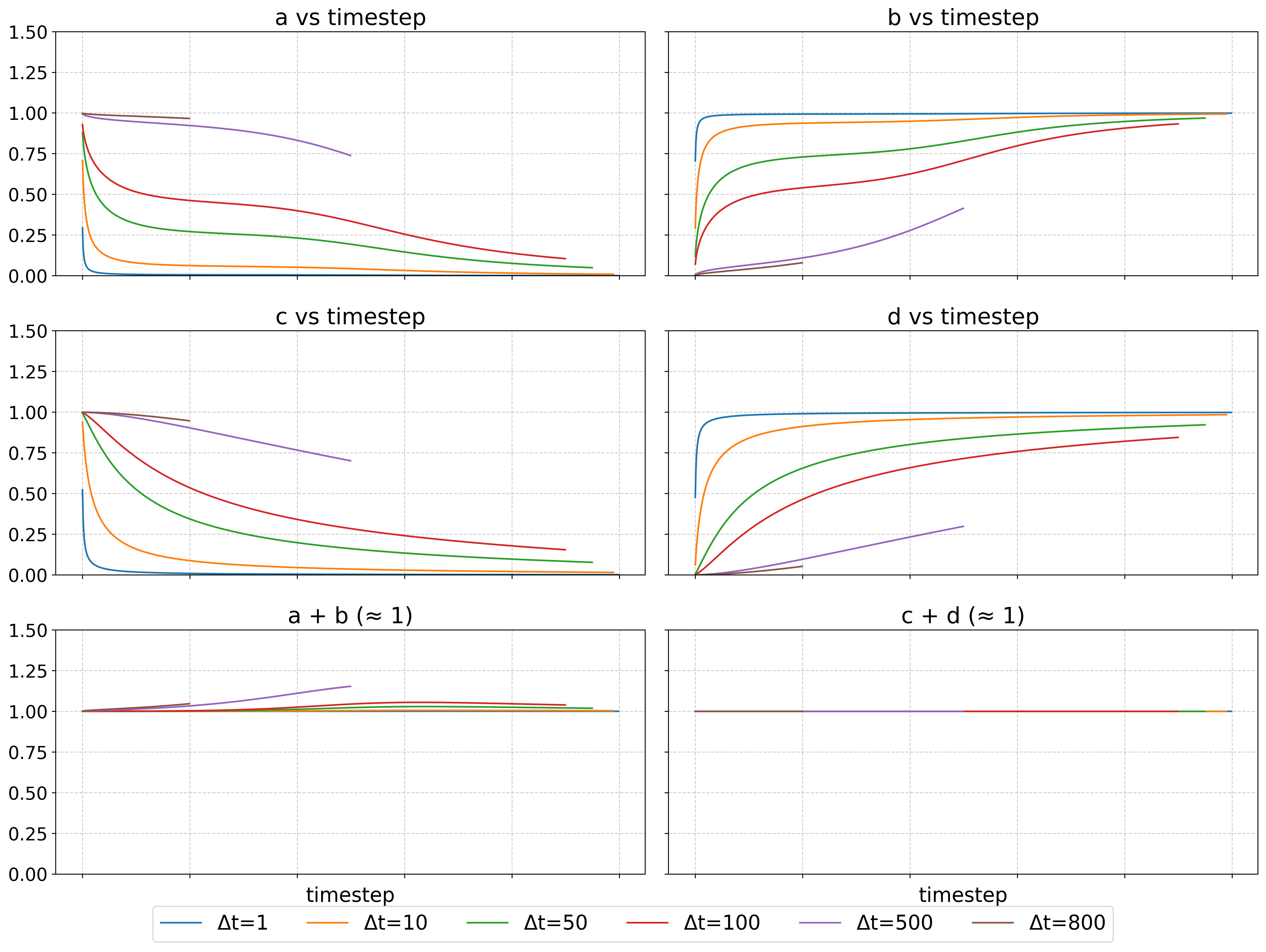}
\caption{
Visualizations of the parameters $a$, $b$, $c$, and $d$ as functions of the diffusion timestep for multiple values of the time interval $\Delta t$. These parameters are derived under the Targetdiff noise scheduling scheme and characterize the interpolation coefficients used in the sampling process. The plots demonstrate how $a$ and $b$ and $c$ and $d$ vary with timestep, as well as the approximate identities $a + b \approx 1$ and $c + d \approx 1$ shown in the bottom row. Different curves correspond to different values of $\Delta t$.
}
\label{noise_analyze}
\end{figure*}

We start from the deterministic accelerated sampling rule and rewrite it in an
explicit interpolation form.
For the continuous coordinate modality, the update from timestep $t$ to
$t-\Delta t$ can be written as:
\begin{equation}
\mathbf{x}_{t-\Delta t}
:= a(t,\Delta t)\hat{\mathbf{x}}_0
+ b(t,\Delta t)\mathbf{x}_t,
\label{eq:ddim_interp}
\end{equation}
where $\hat{\mathbf{x}}_0$ denotes the predicted clean sample at timestep $t$.
An analogous interpolation form holds for the discrete atom-type modality.

Under the standard diffusion parameterization, the interpolation coefficients in
Eq.~\eqref{eq:ddim_interp} admit closed-form expressions:
\begin{align}
a(t,\Delta t)
:&= \sqrt{\overline{\alpha}_{t-\Delta t}}
- \frac{\sqrt{(1-\overline{\alpha}_{t-\Delta t})\overline{\alpha}_t}}
{\sqrt{1-\overline{\alpha}_t}}, \\
b(t,\Delta t)
:&= \frac{\sqrt{1-\overline{\alpha}_{t-\Delta t}}}
{\sqrt{1-\overline{\alpha}_t}}.
\end{align}
For the atom-type modality, the corresponding interpolation takes the form:
\begin{align}
\mathbf{h}_{t-\Delta t}
:&= c(t,\Delta t)\hat{\mathbf{v}}_0
+ d(t,\Delta t)\mathbf{v}_t
+ \frac{1-\overline{\beta}_{t-\Delta t}}{1-\overline{\beta}_t}
\left(\frac{1-\overline{\beta}_t}{\overline{\beta}_t K}\right),
\end{align}
with coefficients:
\begin{align}
c(t,\Delta t)
:&= \overline{\beta}_{t-\Delta t}
- \frac{(1-\overline{\beta}_{t-\Delta t})\overline{\beta}_t}
{1-\overline{\beta}_t}, \\
d(t,\Delta t)
:&= \frac{1-\overline{\beta}_{t-\Delta t}}
{1-\overline{\beta}_t}.
\end{align}

Although the coefficients explicitly depend on the timestep $t$ and step size
$\Delta t$, we empirically observe that for a wide range of $\Delta t$ from Fig~\ref{noise_analyze},
\begin{equation}
a(t,\Delta t) + b(t,\Delta t) \approx 1, \qquad
c(t,\Delta t) + d(t,\Delta t) \approx 1.
\end{equation}

Motivated by the above observation, we eliminate the explicit dependence on
$(t,\Delta t)$ and instead define adaptive interpolation coefficients based on
the discrepancy between the predicted clean sample and the current noisy state.
Specifically, we reformulate the update as:
\begin{equation}
\mathbf{x}_{t-\Delta t}
:= \text{scale} \cdot
\big(a(\hat{\mathbf{x}}_0, \mathbf{x}_t)\hat{\mathbf{x}}_0
+ (1-a(\hat{\mathbf{x}}_0, \mathbf{x}_t))\mathbf{x}_t\big),
\end{equation}
\begin{equation}
\mathbf{h}_{t-\Delta t}
:= c(\hat{\mathbf{v}}_0, \mathbf{v}_t)\hat{\mathbf{v}}_0
+ (1-c(\hat{\mathbf{v}}_0, \mathbf{v}_t))\mathbf{v}_t.
\end{equation}
The adaptive coefficients are defined as:
\begin{align}
a(\hat{\mathbf{x}}_0, \mathbf{x}_t)
:&= \frac{1}{1 + \gamma \log(1 + \|\hat{\mathbf{x}}_0 - \mathbf{x}_t\|_2^2)}, \\
c(\hat{\mathbf{v}}_0, \mathbf{v}_t)
:&= \frac{1}{1 + \eta \log\!\left(1 + \mathrm{KL}
(\hat{\mathbf{v}}_0 \| \mathbf{v}_t)\right)}.
\end{align}

Here, $\|\hat{\mathbf{x}}_0 - \mathbf{x}_t\|_2^2$ and
$\mathrm{KL}(\hat{\mathbf{v}}_0 \| \mathbf{v}_t)$ serve as modality-specific
surrogates for denoising confidence.
By appropriately choosing $\gamma$ and $\eta$, the coordinate and atom-type
modalities can be denoised in a synchronized manner without relying on explicit
time-step scheduling.

\section{Protein Pocket Selection Visualization}

\begin{figure*}[!t]
\centering
\includegraphics[width=0.8\textwidth]{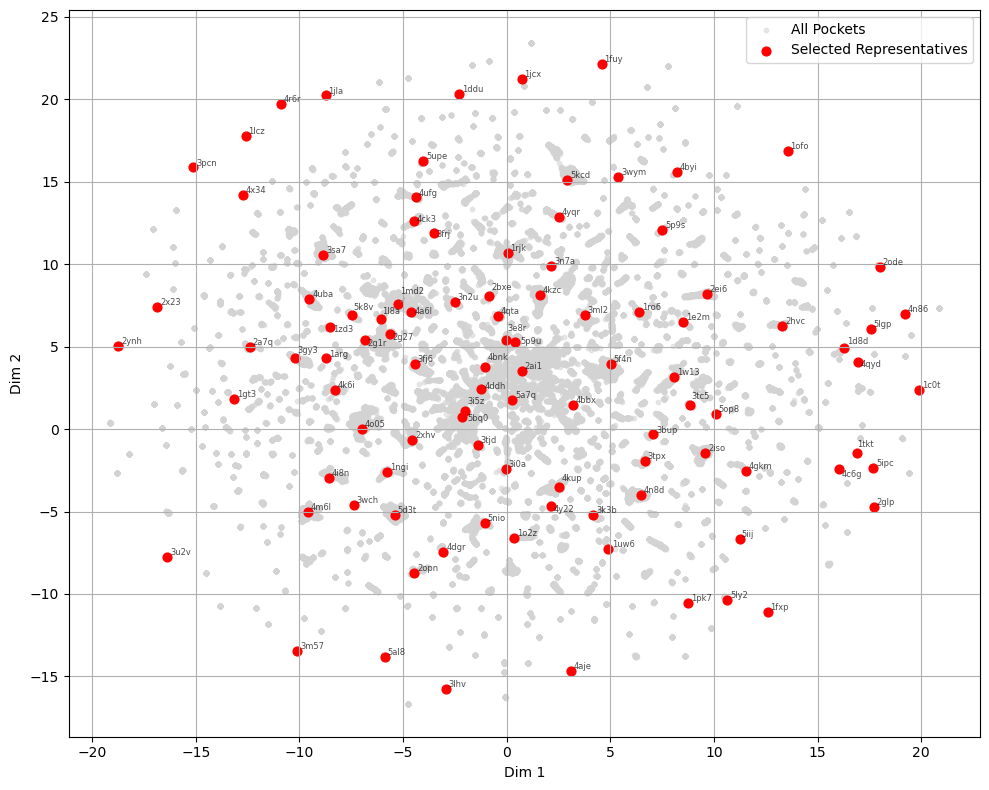}
\caption{
UMAP visualization of protein pocket embeddings. Gray: all pockets; Red:
selected cluster representatives for fine-tuning.  
}
\label{fig:kmeans-vis}
\end{figure*}

To illustrate the selection of protein pockets used for fine-tuning, we
visualize the pocket embeddings in two dimensions using UMAP~\cite{mcinnes2018umap}.
Each point corresponds to a protein pocket in the training dataset.
Gray points indicate all pockets, while red points denote the representatives
closest to the cluster centers obtained via K-means clustering ($k=100$).
These selected representatives form the fine-tuning dataset (see Figure~\ref{fig:kmeans-vis}).

\section{Efficiency Comparison of FTDiff}

\begin{figure}[H]
\centering
\includegraphics[width=0.8\columnwidth]{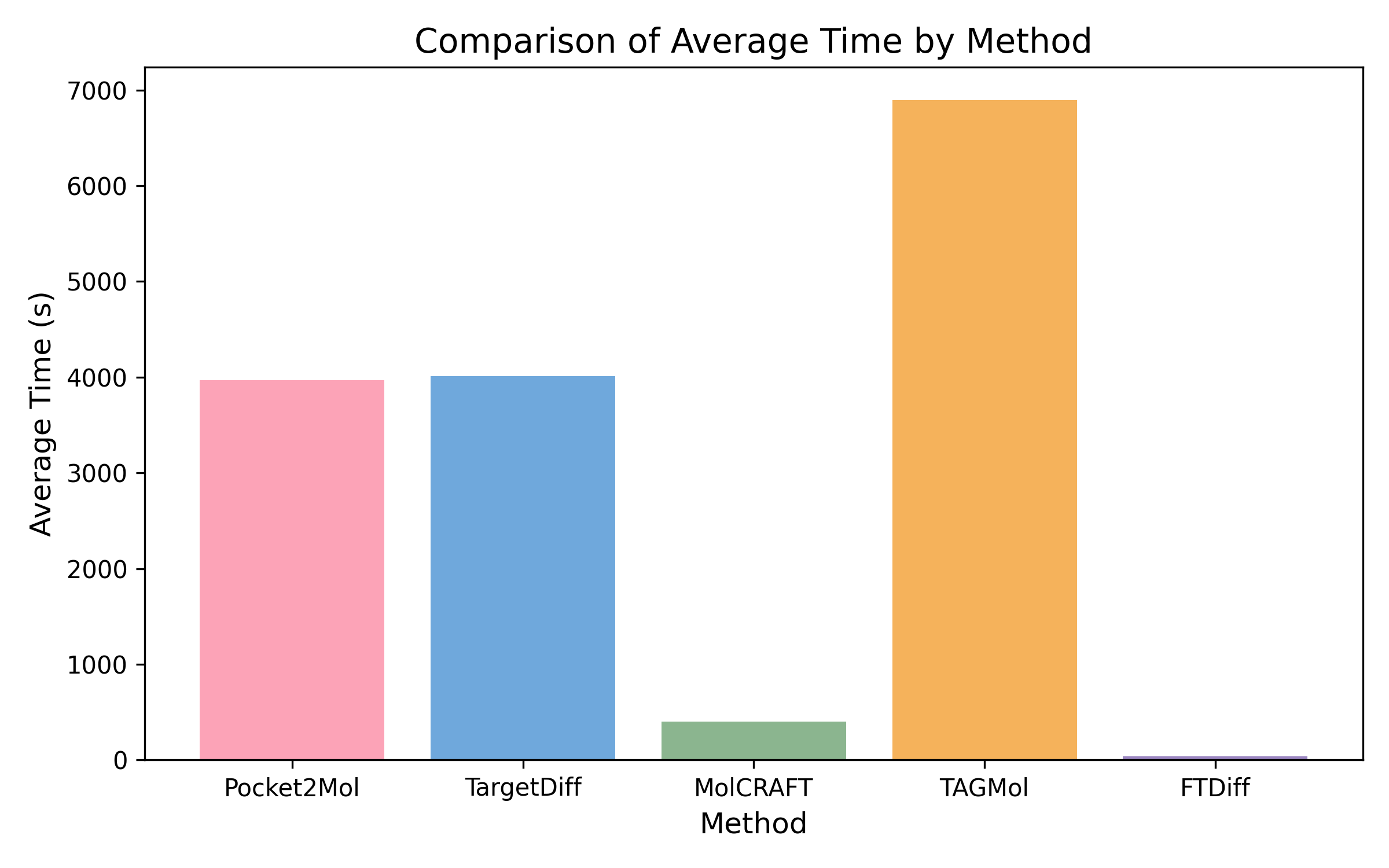}
\caption{
Comparison of average sampling time for different methods. FTDiff demonstrates
substantially faster inference compared with other baselines, while
maintaining molecular quality.}
\label{fig:time_comparison}
\end{figure}

To highlight the computational efficiency of FTDiff, we compare the average
sampling time (in seconds) of several baseline methods and FTDiff on the
same protein pocket test set.  As shown in Figure~\ref{fig:time_comparison}, FTDiff significantly reduces
the runtime compared with other methods. 

\section{Full Algorithm of FTDiff}

Here we provide the complete training procedure of FTDiff, including both
pretraining and fast fine-tuning phases in ~\ref{alg:ftdiff}. This complements Section~3 of the main
paper, where the high-level overview is presented.

\begin{algorithm}[H]
\caption{Training Procedure of FTDiff}
\label{alg:ftdiff}
\begin{algorithmic}[1]
\REQUIRE Pretraining dataset $\mathcal{D}_{\text{pre}} = \{ [\mathbf{x}_0^{(i)}, \mathbf{v}_0^{(i)}], \mathcal{P}^{(i)} \}_{i=1}^{N_{\text{pre}}}$, 
fine-tuning dataset $\mathcal{D}_{\text{ft}} = \{\mathcal{P}^{(j)}\}_{j=1}^{N_{\text{ft}} }$, 
reward model $\mathcal{R}$, denoising network $\phi_\theta$, preference weights $\mathbf{w}$, 
batch size $B$, group size $G$
\ENSURE Finetuned denoising network $\phi_\theta$

\STATE \textbf{Phase I: Diffusion Pretraining}
\FOR{each pretraining iteration}
    \STATE Sample minibatch $\{ [\mathbf{x}_0^{(b)}, \mathbf{v}_0^{(b)}], \mathcal{P}^{(b)} \}_{b=1}^B$ from $\mathcal{D}_{\text{pre}}$
    \STATE Sample timestep $t \sim \mathcal{U}(1, T)$
    \STATE Add noise: $\mathbf{x}_t^{(b)}, \mathbf{v}_t^{(b)} \sim q_t(\cdot \mid \mathbf{x}_0^{(b)}, \mathbf{v}_0^{(b)})$
    \STATE Predict denoised outputs: $[\hat{\mathbf{x}}_0^{(b)}, \hat{\mathbf{v}}_0^{(b)}] = \phi_\theta([\mathbf{x}_t^{(b)}, \mathbf{v}_t^{(b)}], \mathcal{P}^{(b)})$
    \STATE Compute diffusion loss:
    \[
        \mathcal{L}_{\text{diff}} = \sum_{b=1}^B 
        \big\|\hat{\mathbf{x}}_0^{(b)} - \mathbf{x}_0^{(b)}\big\|^2 
        + \text{KL}\big(\mathcal{C}(\mathbf{v}_t^{(b)}) \,\|\, \mathcal{C}(\hat{\mathbf{v}}_{0|t}^{(b)})\big)
    \]
    \STATE Update $\theta \leftarrow \text{Adam}(\nabla_\theta \mathcal{L}_{\text{diff}})$
\ENDFOR

\STATE \textbf{Phase II: GRPO-style Fine-tuning with Time-Free Fast Sampling}
\FOR{each fine-tuning iteration}
    \STATE Sample minibatch of protein pockets $\{\mathcal{P}^{(j)}\}_{j=1}^B$ from $\mathcal{D}_{\text{ft}}$
    \FOR{each protein pocket $\mathcal{P}^{(j)}$}
        \STATE Generate $G$ trajectories $\{ (\mathbf{x}_t^{(g)}, \mathbf{v}_t^{(g)})_{t=1}^T \}_{g=1}^G$ via time-free fast sampling
        \STATE Obtain final denoised molecules $[\mathbf{x}_0^{(g)}, \mathbf{v}_0^{(g)}]$
        \STATE Compute reward $r^{(g)} = \mathcal{R}([\mathbf{x}_0^{(g)}, \mathbf{v}_0^{(g)}], \mathcal{P}^{(j)}, \mathbf{w})$
        \STATE Estimate trajectory-level advantage $A^{(g)}$
        \STATE Store all intermediate states $(\mathbf{x}_t^{(g)}, \mathbf{v}_t^{(g)}, A^{(g)})$ into buffer $\mathcal{B}$
    \ENDFOR

    \FOR{$k = 1$ to $K$}
        \STATE Sample minibatch $(\mathbf{x}_t, \mathbf{v}_t, A)$ from buffer $\mathcal{B}$
        \STATE Predict denoised outputs: $[\hat{\mathbf{x}}_0, \hat{\mathbf{v}}_0] = \phi_\theta([\mathbf{x}_t, \mathbf{v}_t], \mathcal{P})$
        \STATE Compute GRPO-style loss $\mathcal{L}_{\text{DHC}}$ using advantage $A$
        \STATE Update $\theta \leftarrow \text{Adam}(\nabla_\theta \mathcal{L}_{\text{DHC}})$
    \ENDFOR
\ENDFOR

\STATE \textbf{Return} finetuned network $\phi_\theta$
\end{algorithmic}
\end{algorithm}

\end{document}